\documentclass[10pt,twocolumn,letterpaper]{article}

\usepackage{amsthm}
\usepackage{amsmath}
\usepackage{amssymb}
\usepackage{bm}
\usepackage{mathrsfs}
\usepackage[dvips]{graphicx}

\usepackage{algorithm}
\usepackage{algorithmic}

\usepackage{color}

\usepackage{url}

\usepackage{supertabular}

\theoremstyle{definition}

\theoremstyle{remark}

\numberwithin{thm}{section}

\DeclareMathAlphabet{\mathsfsl}{OT1}{cmss}{m}{sl}

\renewcommand{\phi}{\varphi}

\newcommand{\bx}{\boldsymbol{x}}

\newcommand{\by}{\boldsymbol{y}}

\newcommand{\bz}{\boldsymbol{z}}

\def\bx{\boldsymbol{x}}
\def\bga{\boldsymbol{\gamma}}
\def\beps{\boldsymbol{\epsilon}}

\def\bt{\boldsymbol{t}}
\def\bu{\boldsymbol{u}}

\def\b0{\mathbf{0}}

\def\bR{\boldsymbol{R}}

\def\bv{\boldsymbol{v}}

\def\bh{\boldsymbol{h}}

\def\bI{\mathbf{I}}

\usepackage{cvpr}
\usepackage{times}
\usepackage{epsfig}
\usepackage{graphicx}
\usepackage{amsmath}
\usepackage{amssymb}
\usepackage[moderate]{savetrees}
\usepackage{color}
\usepackage{amsthm}
\usepackage{multirow}
\theoremstyle{plain}
\newtheorem{theorem}{Theorem}[section]
\newtheorem{lemma}{Lemma}[section]
\newtheorem{definition}{Definition}[section]
\DeclareMathOperator{\spann}{Span}

\usepackage[pagebackref=true,breaklinks=true,letterpaper=true,colorlinks,bookmarks=false]{hyperref}

\cvprfinalcopy 

\def\cvprPaperID{4239} 

\begin{document}


\title{Estimation of Camera Locations in Highly Corrupted Scenarios:\\
All About that Base, No Shape Trouble}

\author{Yunpeng Shi\\
University of Minnesota\\
{\tt\small shixx517@umn.edu}
\and
Gilad Lerman\\
University of Minnesota\\
{\tt\small lerman@umn.edu}
}

\maketitle
\begin{abstract}
   We propose a strategy for improving camera location estimation in structure from motion. Our setting assumes highly corrupted pairwise directions (i.e., normalized relative location vectors), so there is a clear room for improving current state-of-the-art solutions for this problem. Our strategy identifies severely corrupted pairwise directions by using a geometric consistency condition. It then selects a cleaner set of pairwise directions as a preprocessing step for common solvers. We theoretically guarantee the successful performance of a basic version of our
strategy under a synthetic corruption model. Numerical results on artificial and real data demonstrate the significant improvement obtained by our strategy.
\end{abstract}

\section{Introduction}

The problem of Structure from Motion (SfM), that is, reconstructing 3D structure from 2D images, is critical in computer vision. The common pipeline for 3D reconstruction consists of the following steps: 1.~Matching keypoints among images using SIFT \cite{sift04}; 2.~Computing the essential matrices from the matched image pairs and extracting relative camera rotations \cite{multiviewbook}; 3.~Finding global camera orientations via rotation synchronization and estimating relative camera translations \cite{Nachimson_LS,ChatterjeeG13_rotation, Govindu04_Lie,HartleyAT11_rotation,MartinecP07_rotation, OzyesilSB15_SDR}; 4.~Estimating camera locations from estimated pairwise directions \cite{Nachimson_LS, BrandAT04_LS,  GoldsteinHLVS16_shapekick, Govindu01_LS, Govindu04_Lie,HandLV15,MoulonMM13_Linfty, cvprOzyesilS15,OzyesilSB15_SDR, TronV09_CLS1, TronV14_CLS2}, where a pairwise direction between two cameras is the normalized relative location vector between them; 5.~Recovering the 3D structure using bundle adjustment \cite{bundle99}.  The key for successful 3D recovery is the accurate estimation of camera parameters, including camera locations and orientations. 
These parameters can be misestimated due to erroneous keypoint matching, which results in inaccurate estimates of the essential matrices~\cite{sfmsurvey_2017}.
This paper develops a robust and theoretically-guaranteed strategy for improving camera location estimation from corrupted pairwise directions.
\subsection{Previous Works}
A variety of camera location solvers have been proposed in the past two decades~\cite{sfmsurvey_2017}. The least squares methods \cite{Nachimson_LS, BrandAT04_LS,  Govindu01_LS} are among the earliest solvers. However, these methods are not robust to outliers (namely, maliciously corrupted pairwise directions) and furthermore they typically produce collapsed location estimates. That is, the estimated camera locations are usually clustered around few points. The constrained least squares (CLS) method \cite{TronV09_CLS1,TronV14_CLS2} introduced an anti-collapsed constraint, which makes it more stable to noise but not outliers. The semidefinite relaxation (SDR) solver \cite{OzyesilSB15_SDR} converts the least squares problem into an SDP formulation with a nonconvex anti-collapse constraint. However, it is not outliers-robust, and its computation is challenging even after convex relaxation. Other solvers include the $\ell_\infty$ method \cite{MoulonMM13_Linfty} and the Lie-algebraic averaging method~\cite{Govindu04_Lie}, but the $\ell_\infty$ norm is sensitive to outliers and~\cite{Govindu04_Lie} suffers from convergence to local minima and from sensitivity to outliers.

Recent outlier-robust methods have been proposed for camera location estimation. One class of solvers use outlier detection algorithms as a preprocessing step to improve their subsequent estimator. For the different problem of camera rotation estimation, cycle-consistency constraints were proposed in \cite{MoulonMM13_Linfty,Zach2010} to remove outlying relative orientation measurements. For camera location recovery, the 1DSfM algorithm was proposed in \cite{1dsfm14} for removing outlying pairwise directions. It projects the 3D direction vectors onto 1D, reformulates the cycle-consistency constraints as an ordering problem and solves it using a heuristic combinatorial method. However, its convergence to the global minimum is not guaranteed.  Another class of methods directly solve robust convex optimization problems and include the least unsquared deviations (LUD) algorithm~\cite{cvprOzyesilS15} and the ShapeFit algorithm~\cite{HandLV15}. Exact recovery guarantees under a certain corruption model were established for ShapeFit and LUD in~\cite{HandLV15} and~\cite{LUDrecovery} respectively.
An ADMM-accelerated version of ShapeFit, called ShapeKick, was proposed in~\cite{GoldsteinHLVS16_shapekick}. However, it sacrifices accuracy for speed. A robust formulation for estimating the fundamental matrices was presented in~\cite{SenguptaAGGJSB17}. However, it may suffer from convergence to local minima and requires good initialization.

\subsection{Contribution of This Work}

We propose a novel algorithm for detecting and removing highly corrupted pairwise directions. We use it as a preprocessing step for existing location recovery algorithms.
Our method forms a statistic for any pairwise direction between two given cameras. This statistic estimates the average inconsistency of this pairwise direction with any two pairwise directions associated with an additional camera. This inconsistency is based on the shortest path in $S^2$ between a direction vector and a base of a spherical triangle.
We thus refer to this inconsistency and statistic as All-About-that-Base (AAB). After computing a fast version of the AAB statistic, we remove edges with large statistics and apply a preferable solver. This method is fast and easy to implement, and it can be used as a preprocessing step for any camera location solver. Most importantly, we are able to theoretically guarantee its successful classification on corrupted and uncorrupted edges. We are not aware of any other theoretically-guaranteed algorithm for removing corrupted pairwise direction measurements.
We also present an iterative procedure for improving the AAB statistic, so outliers could be identified more accurately.
Experiments on synthetic and real data demonstrate significant improvement of camera location accuracy by our proposed method.

%

\section{Setting for Camera Location Estimation} \label{sec:math_set}

A mathematical setting for camera location estimation assumes $n$ unknown camera locations $\{\bt_i^*\}_{i\in [n]} \subseteq \mathbb{R}^3$, where $[n]=\{1,2,\dots, n\}$.
The ground truth pairwise direction $\bga^*_{ij}$ between cameras $i$, $j \in [n]$ is defined by
\begin{equation}\label{eq:gammastar}
\bga^*_{ij}=\frac{\bt_i^*-\bt_j^*}{\|\bt_i^*-\bt_j^*\|},
 \end{equation}
where $\|\cdot\|$ denotes the Euclidean norm.
In practice, one often measures a corrupted pairwise direction $\bga_{ij}$  between cameras $i$ and $j$.
The mathematical problem assumes possibly corrupted pairwise measurements $\bga_{ij}\in E$ for some $E\subseteq [n]\times [n]$ and asks to estimate the camera locations $\{\bt_i^*\}_{i\in [n]}$ up to ambiguous translation and scale.
Note that $E$ may not include all the pairs of indices, so that some values can be missing.

In order to establish theoretical guarantees and conduct synthetic data experiments for the AAB procedure, we assume that the true camera locations and corrupted pairwise directions are generated by the following slight modification of the Uniform Corruption Model UC$(n, p, q, \sigma)$~\cite{cvprOzyesilS15}:
Let $V=\{\bt_i^*\}_{i\in [n]}$ be generated by i.i.d.~$N(\boldsymbol 0,\bI_3)$ and let $G(V,E)$ be a graph generated by the Erd\"{o}s-R\'{e}nyi model $G(n,p)$, where $p$ denotes the connection probability among edges. For any $ij \in E$, a corrupted pairwise direction  $\bga_{ij}$ is generated by
\begin{align}
\bga_{ij}=\begin{cases}
\bv_{ij}, & \text{ w.p. } q;\\
\frac{\bga_{ij}^*+\sigma \beps_{ij}}{\|\bga_{ij}^*+\sigma \beps_{ij}\|}, & \text{ w.p. } 1-q,
\end{cases}
\end{align}
where $0<q<1$ is the probability of corruption, $\sigma \geq 0$ is the noise level and $\bv_{ij}$, $\beps_{ij}$ are independently drawn from a uniform distribution on $S^2$. 
The UC model of \cite{cvprOzyesilS15} assumes instead that $\beps_{ij}$ are i.i.d.~$N(\boldsymbol 0,\bI_3)$.
We have noticed similar numerical results for data generated from both models, however, our theory described below is easier to state and verify under the uniform assumption.

\section{Statistics for Corruption Reduction}\label{sec:aab}
We describe a statistic that may distinguish corrupted edges. It uses the geometric notion of cycle-consistency of uncorrupted edges.
Cycle-consistency measures were used in \cite{MoulonMM13_Linfty,1dsfm14,Zach2010} as criteria for outlier removal. For location recovery, the cycle-consistency of 3 vectors $\bga_1$, $\bga_2$, $\bga_3 \in S^2$ refers to the existence of $\lambda_1$, $\lambda_2$, $\lambda_3 > 0$ such that
\begin{equation}\label{eq:cyclecons}
 \lambda_1\bga_1+\lambda_2\bga_2+\lambda_3\bga_3=0.
\end{equation}
One may easily observe that the pairwise directions $\bga_{ij}^*, \bga_{jk}^*$, $\bga_{ki}^*$  are cycle-consistent by substituting in \eqref{eq:cyclecons} $\lambda_{ij}=\|\bt_i^*-\bt_j^*\|$, $\lambda_{jk}=\|\bt_j^*-\bt_k^*\|$ and $\lambda_{ki}=\|\bt_k^*-\bt_i^*\|$.
However, if any of the three vectors is randomly corrupted, the consistency constraint is most probably violated. Thus, we may define a certain cycle-inconsistency measure that indicates the underlying corruption level.

Section \ref{sec:aabi} describes a basic measure of inconsistency of a given pairwise direction 
with respect to 2 other pairwise directions, where the 3 directions result from 3 unknown locations. It is referred to as the AAB inconsistency. A formula for efficiently computing it is proposed at the end of this section. Section \ref{sec:naab} uses these inconsistencies to define the naive AAB statistic of a given pairwise direction that is used to remove corrupted edges.
Section \ref{sec:iraab} discusses the iteratively reweighted AAB (IR-AAB) statistic, which aims to further improve the accuracy of naive AAB in removing corrupted edges. At last, Section \ref{sec:num_consider} discusses some issues regarding practical implementation of naive AAB and IR-AAB.

\subsection{AAB Inconsistency and Formula}\label{sec:aabi}
\begin{figure}[!htbp]
\begin{center}
   \includegraphics[width=0.65\linewidth]{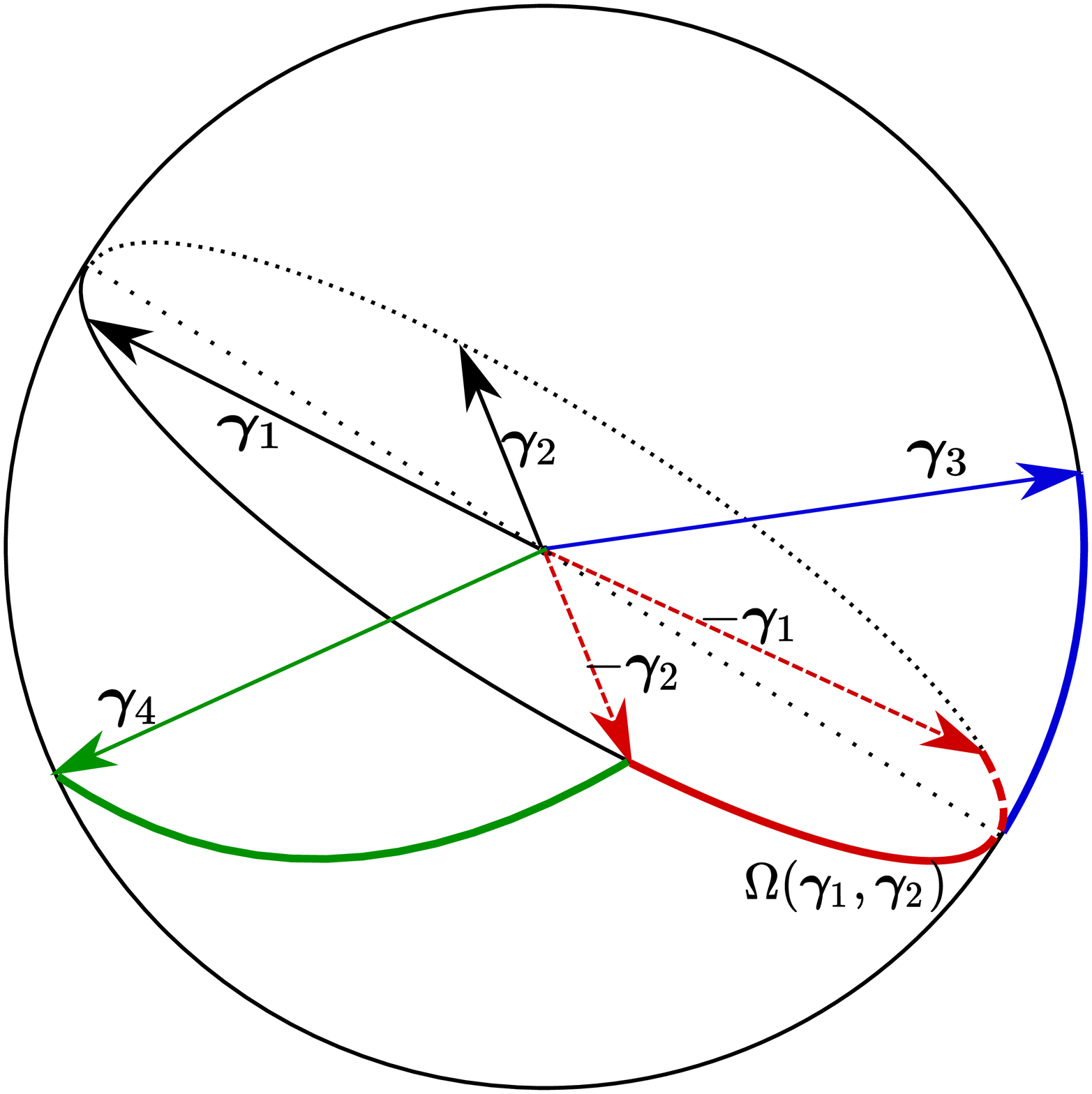}
\end{center}
   \caption{Clarification of the AAB Inconsistency. The red arc is the cycle-consistency region $\Omega(\bga_1,\bga_2)$. Indeed, it follows from \eqref{eq:cyclecons} that the points in $\Omega(\bga_1,\bga_2)$ are linear combinations in $S^2$ with positive coefficients of $-\bga_1$ and $-\bga_2$.   The AAB inconsistency $I_{AAB}(\bga_3;\bga_1,\bga_2)$ is the distance in $S^2$ of $\bga_3$ from $\Omega(\bga_1,\bga_2)$ and is the length of the blue arc. Similarly, $I_{AAB}(\bga_4;\bga_1,\bga_2)$ is the length of the green arc. \label{fig:s2}}
\end{figure}

We define the cycle-consistency region of $\bga_1$, $\bga_2 \in S^2$  as $\Omega(\bga_1,\bga_2)=\{\bga\in S^2: \bga_1, \bga_2, \bga \text{ are cycle-consistent}\}$. We denote by $d_g$ the great-circle distance, i.e., the length of the shortest path on $S^2$. The AAB inconsistency of $\bga_3 \in S^2$ with respect to $\bga_1$ and $\bga_2$ is defined by
\begin{align}
I_{AAB}(\bga_3;\bga_1,\bga_2)&=d_g(\bga_3, \Omega(\bga_1,\bga_2))\nonumber\\
&=\min_{\bga\in \Omega(\bga_1,\bga_2)}d_g(\bga_3, \bga).\label{eq:dg}
\end{align}
Figure \ref{fig:s2} shows that $I_{AAB}(\bga_3;\bga_1,\bga_2)$ is the smallest angle needed to rotate $\bga_3$ so that $\bga_1, \bga_2, \bga_3$ are cycle-consistent.

The following 
formula for computing the AAB inconsistency is crucial for efficient implementation of the algorithms described below.
Its proof appears in Appendix~\ref{ap:aabf}.
For $\bga_1$, $\bga_2$, $\bga_3 \in S^2$, $x=\bga_1^T\bga_3$, $y=\bga_2^T\bga_3$, $z=\bga_1^T\bga_2$ and $a = I(x<yz)\cdot I(y<xz)$, where $I$ is the indicator function,
\begin{multline}\label{eq:aabf}
I_{AAB}(\bga_3;\bga_1,\bga_2)=\\
\cos^{-1}\left(a\cdot \frac{x^2+y^2-2xyz}{1-z^2}+(a-1)\min(x,y)\right).
\end{multline}

\subsection{The Naive AAB Statistic}\label{sec:naab}

We initially define the naive AAB statistic of an edge $ij \in E$ as the average of the AAB inconsistencies $I_{AAB}(\bga_{ij};\bga_{jk},\bga_{ki})$ over the set $C_{ij}=\{k\in[n]: ik\in E \text{ and } jk\in E\}$.
That is,
\begin{equation}
S_{AAB}^{\text{initial}}(ij)=\frac{1}{|C_{ij}|}\sum_{k\in C_{ij}}I_{AAB}(\bga_{ij};\bga_{jk},\bga_{ki}).
\end{equation}
We use it as an indication for the corruption level of $\bga_{ij}$ and thus remove the edges with largest AAB statistics.
Note that the AAB formula in \eqref{eq:aabf} enables computation of the naive AAB statistic through vectorization instead of using a loop, and thus allows efficient coding in programming languages with an effective linear algebra toolbox. However, the average over $C_{ij}$ can be costly and we thus advocate using a small random sample from $C_{ij}$ of size $s$, where the default value of $s$ is 50. We summarize this basic procedure of computing the AAB statistic, $S_{AAB}^{(0)}$, in Algorithm \ref{alg:naab}.
\begin{algorithm}[!htbp]
\caption{Computation of the Naive AAB Statistic}
\begin{algorithmic}
\REQUIRE $\{\bga_{ij}\}_{ij\in E}$: pairwise directions, $s$: number of samples
\FOR{each $ij\in E$}
\STATE $S_{ij}$ = $s$ random samples with replacement from $C_{ij}$
\STATE $S_{AAB}^{(0)}(ij)=\frac{1}{s}\sum_{k\in S_{ij}}I_{AAB}(\bga_{ij};\bga_{jk},\bga_{ki})$
\ENDFOR
\ENSURE Naive AAB statistic $\left\{S_{AAB}^{(0)}(ij)\right\}_{ij\in E}$
\end{algorithmic}\label{alg:naab}
\end{algorithm}

\subsection{Iteratively Reweighted AAB}\label{sec:iraab}

The naive AAB statistic may suffer from unreliable AAB inconsistencies when the corruption level $q$ is high. Specifically, for an uncorrupted direction $\bga_{ij}$, its AAB inconsistency with respect to $\bga_{jk} $ and $\bga_{ki}$ can be unreasonably high if either $\bga_{jk}$ or $\bga_{ki}$ is severely corrupted. Moreover, if many adjacent edges of $ij$ are corrupted, then the naive AAB statistic of this edge may not accurately measure its corruption level. The main issue is not the misleading effect of neighboring  edges, but the fact that only such edges are considered and relevant information from other edges is not incorporated. To overcome this issue, the iteratively reweighted AAB (IR-AAB) statistic computes a weighted mean of AAB inconsistencies and iteratively updates these weights. This results in propagation of global information from other non-neighboring edges to edge $ij$.

\begin{figure}[!htbp]
\begin{center}
   \includegraphics[width=1\linewidth]{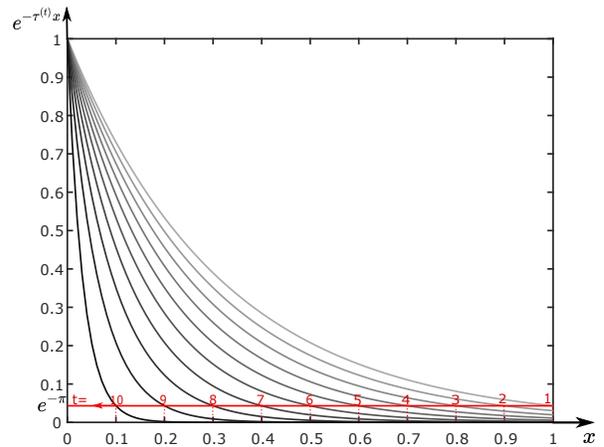}
\end{center}
   \caption{Demonstration of the reweighting function $\exp(-\tau^{(t)}x)$ used in IR-AAB. Here, $t\in [10]$ and the rate of decrease is $\tau^{(t)}=\pi/(1.1-0.1t)$, which increases with $t$. The labels on the x-axis are of the points $x_t=1.1-0.1t$, $1\leq t\leq 10$. At each iteration $t$, $\exp(-\tau^{(t)}x)<e^{-\pi}\approx 0.04$ for $x>x_t$. The red line separates for each curve the values in $[0,x_t]$ and $[x_t,1]$. Therefore,  $\exp(-\tau^{(t)}x)$ gives little weight to points in $[x_t,1]$.\label{fig:exp}}
\end{figure}

Initially, the IR-AAB procedure computes the naive AAB statistic. 
The reweighting strategy of IR-AAB tries to reduce the weights of $I_{AAB}(\bga_{ij};\bga_{jk},\bga_{ki})$ when either $ki$ or $kj$ are highly corrupted. In order to do this, at each iteration the AAB inconsistencies $I_{AAB}(\bga_{ij};\bga_{jk},\bga_{ki})$ involving suspicious edges are penalized by the reweighting function $\exp(-\tau^{(t)}x)$. The number $x$ is the maximal value of the reweighted AAB statistics computed in previous iteration for edges $ik$ and $kj$. The parameter $\tau^{(t)}$ increases iteratively and depends on the initial maximal and minimal values of inconsistencies, denoted by $M$ and $m$.
Figure \ref{fig:exp} illustrates the reweighting functions with  $M=1$, $m=0$ and $10$ iterations.
The use of slowly-decreasing reweighting functions in the first iterations ensures that only the most unreliable AAB inconsistencies are ignored. As the data is iteratively purified, the AAB inconsistencies involving ``good" edges are weighted more and more. We remark that increasing $\tau^{(t)}$ corresponds to focusing more on ``good" edges and ignoring more ``suspicious" edges. The details of computing the IR-AAB statistic are described in Algorithm \ref{alg:iraab}.
\begin{algorithm}[!htbp]
\caption{Computation of the IR-AAB statistic}
\begin{algorithmic}
\REQUIRE $\{\bga_{ij}\}_{ij\in E}$: pairwise directions, $s$: number of samples, $T$: number of iterations
\STATE Compute $S_{ij}$, $S_{AAB}^{(0)}(ij)$  $ \forall ij\in E$ by Algorithm~\ref{alg:naab}
\STATE $M=\max_{ij\in E,k\in S_{ij}}I_{AAB}(\bga_{ij};\bga_{jk},\bga_{ki})$
\STATE $m=\min_{ij\in E,k\in S_{ij}}I_{AAB}(\bga_{ij};\bga_{jk},\bga_{ki})$
\STATE $L=(M-m)/T$
\FOR {$t=1:T$}
\STATE $\tau^{(t)}= \pi/M$
\STATE $M=M-L$
\FOR {$ij\in E$ and $k\in S_{ij}$}
\STATE $w_{ij,k}^{(t)}= \exp\left(-\tau^{(t)}\max\left\{S_{AAB}^{(t-1)}(ki),S_{AAB}^{(t-1)}(jk)\right\}\right)$
\STATE $w_{ij,k}^{(t)}= w_{ij,k}^{(t)}/\sum_{k\in S_{ij}}{w_{ij,k}^{(t)}}$
 \STATE  $S_{AAB}^{(t)}(ij)= \sum_{k\in S_{ij}}w_{ij,k}^{(t)}I_{AAB}(\bga_{ij};\bga_{jk},\bga_{ki})$
\ENDFOR
\ENDFOR
\ENSURE IR-AAB statistic: $\left\{S_{AAB}^{(T)}(ij)\right\}_{ij\in E}$
\end{algorithmic}\label{alg:iraab}
\end{algorithm}

Note that IR-AAB alternatively updates the weights using the AAB statistics and then updates the AAB statistics using the new weights. This way better weights can reduce the effect of highly corrupted edges so that the updated AAB statistics measures more accurately the corruption level of edges. Similarly, better estimates of the corruption level by the AAB statistics provide more accurate weights, which emphasize the more relevant edges.
In the special practical case of repetitive patterns (e.g., due to identical windows), this procedure can help in identifying corrupted edges that are self-consistent with each other.

At last we comment that the failure mode for any AAB procedure is when there are no outliers, so the task of identifying corruptions is ill-posed. This can also happen when the noise magnitude is enormous and outliers are not distinguishable.

\subsection{Numerical Considerations}\label{sec:num_consider}
As mentioned earlier, implementations for naive AAB and IR-AAB may avoid loops and use instead vectorization due to the AAB formula. An efficient Matlab code will be provided in the future supplemental webpage. For naive AAB and IR-AAB we recommend using $s=50$ as default, and we applied this value in all of our experiments. For IR-AAB we recommend and implement the default value $T=10$.

We note that the computational complexity of naive AAB is $O(s\cdot|E|)$, where $|E|$ is the number of edges. In general, for dense graphs the complexity is $O(s\cdot n^2)$, but for sparser graphs the complexity decreases, e.g., for sparse Erd\"{o}s-R\'{e}nyi graphs with $p \ll 1$, the complexity is $O_P(s \cdot p \cdot n^2)$ since $\mathbb{E}[|E|]= n(n-1)p/2$.
The computational complexity of IR-AAB is also $O(s\cdot|E|)$. While IR-AAB is iterated $T=10$ times, its main computation is due to the initial application of naive AAB, which requires the computation of the AAB inconsistencies. On the other hand the weight computations in the subsequent iterations is much cheaper. Therefore in practice, the computational complexity of naive AAB and IR-AAB are truly comparable.

For synthetic data, we demonstrate in Section \ref{sec:syn} that a threshold on the naive AAB and IR-AAB statistics can be chosen by their corresponding histograms. We also demonstrate performance with differently chosen thresholds via ROC curves. The histograms of real data are not so simple, and thus in this case we keep half of the edges with the lowest values of the corresponding statistic. We have noticed that the less edges we keep the higher accuracy we obtain for location estimation. However, extremely low threshold results in limited number of camera locations. Demonstrations of other thresholds appear in Appendix~\ref{ap:real}.

\section{Theoretical Guarantees for Outliers Removal}
\label{sec:theory}
We show that the naive AAB statistic can be used for near-perfect separation of corrupted and uncorrupted edges.
Given pairwise directions generated on an edge set $E$ by the uniform corruption model, we denote by $E_g$ the uncorrupted edges, namely, all edges $ij \in E$ such that $\bga_{ij}=\bga^*_{ij}$. We denote the rest of edges in $E$ by $E_b$.
The theorem below states that under the uniform corruption model with sufficiently small corruption probability and noise level, the naive AAB statistic is able to perfectly separate $E_g$ as well as a large portion of $E_b$.
\begin{theorem}\label{thm:main}
There exist absolute positive constants $C_0, C$ such that for any $\epsilon\in [0,1]$ and for pairwise directions randomly generated by the uniform corruption model UC($n,p,q,\sigma$) with $n=\Omega(1/pq\epsilon)$, $np^2(1-q)^2\geq C_0\log n$ and  $q+\sigma<C\epsilon/\sqrt{\log n}$, there exists a set $E'\subseteq E_b$ such that $|E'|\geq (1-\epsilon)|E_b|$ and with probability 
$1-O(n^{-5})$,
\begin{equation}\label{eq:thm}
\min_{ij\in E'}\mathbb{E}[S^{(0)}_{AAB}(ij)]> \max_{ij\in E_g}\mathbb{E}[S^{(0)}_{AAB}(ij)].
\end{equation}
\end{theorem}

The theorem can be extended to other synthetic models. For instance, the assumption in the UC model that the locations are sampled from a Gaussian distribution can be generalized to any distribution that generates ``c-well-distributed locations", which are explained in Section \ref{sec:proof_preliminary}.   One can show that a compactly supported distribution with continuous and positive density satisfies this criterion with an absolute constant $c$ (unlike the Gaussian case) and consequently the theorem may have the weaker assumption: $q+\sigma<C\epsilon$.
The uniform noise assumption in the UC model of this paper can be directly extended to any compactly supported distribution.
For Gaussian noise, one needs to slightly modify the theorem so the RHS of \eqref{eq:thm} is maximized over a sufficiently large subset of $E_g$ (similarly to the LHS w.r.t.~$E_b$).

\subsection{Proof of Theorem \ref{thm:main}}\label{sec:pf}

After reviewing preliminary results and notation in Section \ref{sec:proof_preliminary},
Section \ref{sec:key} describes the main part of the proof. It starts with stating two essential bounds: An upper bound on the expectation of $S^{(0)}_{AAB}(ij)$ when $ij\in E_g$
and a lower probabilistic bound on the expectation of $S^{(0)}_{AAB}(ij)$ when $ij\in E_b$.
The upper bound is stated in \eqref{eq:Egupper} and later proved in Section \ref{sec:eq1}.
The lower bound is stated in \eqref{eq:Eblower} and later proved in proved in Section \ref{sec:eq2}.
While the upper bound is uniform over $ij\in E_g$, the lower bound depends on the corruption level of each edge $ij\in E_b$. However, there is an absolute bound which holds within a large subset of $E_b$.
We show that the uniform upper bound is lower than the absolute lower bound and thus conclude that with high probability the expected values of $S^{(0)}_{AAB}(ij)$ when $ij\in E_g$ are separated from the expected values of $S^{(0)}_{AAB}(ij)$ when $ij$ is in a large subset of $E_b$.

\subsubsection{Preliminaries}
\label{sec:proof_preliminary}
We first summarize some properties of the AAB inconsistency:\\
(i) $I_{AAB}(\bga_3;\bga_1,\bga_2)\in [0,\pi]$\  $\forall$ $\bga_1,\bga_2,\bga_3\in S^2$.\\
(ii) $I_{AAB}(\bga_3;\bga_1,\bga_2)=0$ iff $\bga_1,\bga_2,\bga_3$ are cycle-consistent.\\
(iii) The AAB inconsistency is rotation-invariant. That is, for any rotation $\bR$: $I_{AAB}(\bga_3;\bga_1,\bga_2)=I_{AAB}(\bR\bga_3;\bR\bga_1,\bR\bga_2)$.

We denote by $U(S^2)$ the uniform distribution on $S^2$ and define $Z:=I_{AAB}(\bz;\bx,\by)$,  where  $\bx$, $\by$, $\bz$ i.i.d.~$\sim U(S^2)$.
For $x\in [0,\pi]$, let $f(x):=\mathbb{E}[I_{AAB}(\bv_2(x);\bv_1,\bv)|\bv\sim U(S^2)]$, where $\bv_1=(-1,0,0)^T$ and $\bv_2(x)=(\cos x,\sin x,0)^T$. The following property is proved in Appendix~\ref{ap:fx}.
\begin{lemma}\label{thm:f}
If $x\in [0,\pi]$, then $f(x)=\frac12(x+\sin x)$.
\end{lemma}

We will use the following definition and Lemma of \cite{HandLV15}.
\begin{definition}[Definition 2 of \cite{HandLV15}]\label{def:cdist}
Let $G=G(V,E)$ be a graph with vertices $V = \{\bt_i\}_{i=1}^n\subseteq \mathbb{R}^3$. For $\bx$, $\by\in \mathbb{R}^3$, $c>0$ and $A\subseteq V$, we say that $A$ is $c$-well-distributed
with respect to $(\bx, \by)$ if the following holds for any $\bh \in \mathbb{R}^3$:
\begin{equation}
\frac{1}{|A|}\sum_{\bt\in A} \|P_{\spann\{\bt-\bx,\bt-\by\}^\perp}(\bh)\| \geq c · \|P_{(\bx-\by)^\perp}(\bh)\|.
\end{equation}
We say that $V$  is $c$-well-distributed along $G$ if for all distinct $1\leq i, j \leq n$, the set $S_{ij} = \{\bt_k \in V:
ik, jk \in E(G)\}$ is $c$-well-distributed with respect to $(\bt_i, \bt_j)$.
\end{definition}
\begin{lemma}[Lemma 18 of \cite{HandLV15}]\label{thm:cdist}
Assume that $V=\{\bt_i\}_{i=1}^n$ is i.i.d.~generated by $N(\boldsymbol 0, \bI_3)$ and the graph $G(V,E)$ is generated by the Erd\"{o}s-R\'{e}nyi model $G(n,p)$. There exist absolute positive constants $C_0, C_1$ such that if $np^2\geq C_0\log n$, then with probability $1-n^{-5}$, the set $V$ is $C_1/\sqrt{\log n}$-well-distributed along $G$.
\end{lemma}

\subsubsection{The Main Part of the Proof}\label{sec:key}
Let $e_{ij}:=\measuredangle(\bga_{ij},\bga_{ij}^*)$ denote the corruption level of edge $ij\in E$. We later prove in Sections \ref{sec:eq1} and \ref{sec:eq2} respectively the following  two inequalities. The first one holds for any fixed $ij\in E$:
\begin{equation}
\mathbb{E}[S^{(0)}_{AAB}(ij)|ij\in E_g]
\leq \pi\sigma(1-q)^2+\pi q(1-q)+q^2\mathbb{E}[Z].\label{eq:Egupper}
\end{equation}
The second one holds with probability $1-n^{-5}$ for all $ij\in E$:
\begin{align}
\mathbb{E}[S^{(0)}_{AAB}(ij)|ij\in E_b]\geq &(1-q)^2\left[\frac{C'}{\sqrt{\log n}}\min(e_{ij}, \pi-e_{ij})-\frac{\pi}{2}\sigma\right]\nonumber\\
&+q^2\mathbb{E}[Z].\label{eq:Eblower}
\end{align}

We conclude the proof by assuming these inequalities. Recall that there exists an absolute constant $C$ such that
\begin{equation}\label{eq:assump}
q+\sigma<\frac{C\epsilon}{\sqrt{\log n}}.
\end{equation}
Multiplying both sides of \eqref{eq:assump} by $3\pi(1-q)^2/2$, noting that for $n$ sufficiently large $1-q\geq 1-C\epsilon/\sqrt{\log n}> 2/3$ and thus
$3 q (1-q)^2 /2> q(1-q)$ and setting $C'= 6C$ yield
\begin{equation}\label{eq:relative}
\pi \left[q(1-q)+\frac{3}{2}\sigma (1-q)^2\right]<(1-q)^2\frac{C'}{\sqrt{\log n}}\cdot\frac{\pi\epsilon}{4}.
\end{equation}
Clearly \eqref{eq:relative} can be rewritten as
\begin{equation}\label{eq:relative1}
\pi\sigma(1-q)^2+\pi q(1-q)<(1-q)^2\left[\frac{C'}{\sqrt{\log n}}\cdot\frac{\pi\epsilon}{4}-\frac{\pi}{2}\sigma\right].
\end{equation}
Combining \eqref{eq:Egupper}, \eqref{eq:Eblower} and \eqref{eq:relative1} results in
\begin{equation}
\max_{ij\in E_g}\mathbb{E}[S^{(0)}_{AAB}(ij)]<\min_{\substack{ij\in E_b\\ \min(e_{ij}, \pi-e_{ij})>\frac{\pi\epsilon}{4}}}\mathbb{E}[S_{AAB}^{(0)}(ij)].
\end{equation}

Let $E'=\{ij\in E_b:\min(e_{ij}, \pi-e_{ij})>\pi\epsilon/4\}$. Since $e_{ij}$ is i.i.d.$\sim U[0,\pi]$, $X_{ij}:=I(ij\notin E')$ is a Bernoulli random variable with mean $\mu=\epsilon/2$. Applying Chernoff bound~\cite{chernoff} yields
\begin{equation}
\Pr\Big(\sum_{ij\in E_b} X_{ij}>2|E_b|\mu\Big)<\exp\left(-\Omega\left(|E_b|\mu\right)\right).
\end{equation}
That is, with probability $1-\exp\left(-\Omega(n^2pq\epsilon)\right)$,
$|E'|>(1-\epsilon)|E_b|$.
Since $n=\Omega(1/pq\epsilon)$ this probability is sufficiently high. Thus, Theorem \ref{thm:main} is concluded if \eqref{eq:Egupper} and \eqref{eq:Eblower} are correct.

\subsubsection{Proof of Inequality \eqref{eq:Egupper}}\label{sec:eq1}
We investigate the distribution of $I_{AAB}(\bga_{ij};\bga_{jk},\bga_{ki})$ for fixed $ij\in E_g$ and $k\in C_{ij}$ in the following 3 complementary cases:\\
\textbf{Case 1}. $jk$, $ki\in E_g$.\\
In this case, $\bga_{ij}=\bga_{ij}^*+ \bv_{ij}$, $\bga_{jk}=\bga_{jk}^*+ \bv_{jk}$ and $\bga_{ki}=\bga_{ki}^*+ \bv_{ki}$, where $\bv_{ij}=(\bga_{ij}^*+\sigma\beps_{ij})/\|\bga_{ij}^*+\sigma\beps_{ij}\|-\bga_{ij}^*$, $\beps_{ij}\sim U(S^2)$ and $\bv_{jk}$ and $\bv_{ki}$ are defined in the same way. We note that if $\sigma=0$, then the AAB inconsistency is $0$ in the current case. If $\sigma>0$, then since $\|\beps_{ij}\|=1$ the AAB inconsistency is bounded as follows:
\begin{align}
X_{ij}^g(k):=&I_{AAB}(\bga_{ij};\bga_{jk},\bga_{ki})\nonumber\\
=&d_g\left(\bga_{ij}^*+\bv_{ij},\Omega(\bga_{jk}^*+\bv_{jk},\bga_{ki}^*+\bv_{ki})\right)\nonumber\\
\leq&d_g\left(\bga_{ij}^*+\bv_{ij},\bga_{ij}^*\right)+d_g\left(\bga_{ij}^*,\Omega(\bga_{jk}^*+\bv_{jk},\bga_{ki}^*+\bv_{ki})\right)\nonumber\\
\leq& d_g\left(\bga_{ij}^*+\bv_{ij},\bga_{ij}^*\right)+d_g\left(\bga_{ij}^*,\Omega(\bga_{jk}^*,\bga_{ki}^*)\right)\nonumber\\
&+\max\left(d_g(\bga_{jk}^*, \bga_{jk}^*+\bv_{jk}),d_g(\bga_{ki}^*, \bga_{ki}^*+\bv_{ki})\right)\nonumber\\
\leq& \frac{\pi}{2}\sigma+0+\frac{\pi}{2}\sigma =\pi\sigma.\label{eq:xg}
\end{align}
\\
\textbf{Case 2}. Either $jk\in E_g$ or $ki \in E_g$, but not both in $E_g$.\\
We assume WLOG that $jk\in E_g$ and $ki\in E_b$. According to the uniform corruption model, $\bga_{ki}\sim U(S^2)$, $\bga_{jk}=\bga_{jk}^*+\bv_{jk}$, $\bga_{ij}=\bga_{ij}^*+\bv_{ij}$. For any indices $ijk$, let $\theta_{ijk}$ denotes the angle between $\bga_{ij}$ and $\bga_{kj}$. By choosing appropriate rotation matrix $\bR$,
\begin{align}
Y^g_{ij}(k)&:=I_{AAB}(\bga_{ij};\bga_{jk},\bga_{ki})=I_{AAB}(\bR\bga_{ij};\bR\bga_{jk},\bR\bga_{ki})\nonumber\\
&=I_{AAB}(\bv_2(\theta_{ijk});\bv_1,\bv),
\end{align}
where $\bv_1$ and $\bv_2(\theta_{ijk})$ were defined in Section \ref{sec:proof_preliminary} and $\bv\sim U(S^2)$. Lemma \ref{thm:f} and the fact that $f(x)\in [0,\pi/2]$ for $x\in [0,\pi]$ imply the inequality
\begin{align}\label{eq:yg}
\mathbb{E}[Y^g_{ij}(k)]=\mathbb{E}_{\theta_{ijk}}[f(\theta_{ijk})]\leq \frac{\pi}{2}.
\end{align}
\textbf{Case 3}. $jk$, $ki\in E_b$\\
Let $Z_{ij}^g(k)$ be defined as follows with distribution equivalent formulations that use an arbitrary rotation $\bR$  and $\bx$, $\by\sim U(S^2)$:
\begin{align}
Z_{ij}^g(k):=I_{AAB}(\bga_{ij};\bga_{jk},\bga_{ki})&\overset{d}{=}I_{AAB}(\bR\bga_{ij};\bR\bga_{jk},\bR\bga_{ki})\nonumber\\
&\overset{d}{=}I_{AAB}(\bR\bga_{ij};\bx,\by).
\end{align}
Since $\bR$ is arbitrary, $Z_{ij}^g(k)$ is independent of $\bga_{ij}$ and for $\bz\sim U(S^2)$
\begin{equation}\label{eq:zg}
Z_{ij}^g(k):=I_{AAB}(\bga_{ij};\bga_{jk},\bga_{ki})\overset{d}{=}I_{AAB}(\bz;\bx,\by)=Z.
\end{equation}

At last, combining \eqref{eq:xg}, \eqref{eq:yg} and \eqref{eq:zg} with probabilities $(1-q)^2$, $2q(1-q)$ and $q^2$ for each case respectively yields \eqref{eq:Egupper}.
\subsubsection{Proof of Inequality \eqref{eq:Eblower}}\label{sec:eq2}
We investigate the distribution of $I_{AAB}(\bga_{ij};\bga_{jk},\bga_{ki})$ for fixed $ij\in E_b$ and $k\in C_{ij}$ in the following 3 complementary cases:\\
\textbf{Case 1}. $jk, ki\in E_g$.
Observe that
\begin{multline}
I_{AAB}(\bga_{ij};\bga_{jk}^*,\bga_{ki}^*)=\min_{v\in \Omega(\bga_{jk}^*,\bga_{ki}^*)}d_g(\bga_{ij}, v)\\
\geq \min_{v\in \spann\{\bga_{jk}^*,\bga_{ki}^*\}}d_g(\bga_{ij}, v)
\geq \min_{v\in \spann\{\bga_{jk}^*,\bga_{ki}^*\}}\|\bga_{ij}-v\|\\
=\|P_{\spann\{\bt_k^*-\bt_i^*,\bt_k^*-\bt_j^*\}^\perp}(\bga_{ij})\|
\end{multline}
and
\begin{align}
X_{ij}^b(k):=&I_{AAB}(\bga_{ij};\bga_{jk},\bga_{ki})\nonumber\\
=&d_g\left(\bga_{ij},\Omega(\bga_{jk}^*+\bv_{jk},\bga_{ki}^*+\bv_{ki})\right)\nonumber\\
\geq& d_g\left(\bga_{ij},\Omega(\bga_{jk}^*,\bga_{ki}^*)\right)\nonumber\\
&-\max\left(d_g(\bga_{jk}^*, \bga_{jk}^*+\bv_{jk}),d_g(\bga_{ki}^*, \bga_{ki}^*+\bv_{ki})\right)\nonumber\\
=& I_{AAB}(\bga_{ij};\bga_{jk}^*,\bga_{ki}^*)\nonumber\\
&-\max\left(d_g(\bga_{jk}^*, \bga_{jk}^*+\bv_{jk}),d_g(\bga_{ki}^*, \bga_{ki}^*+\bv_{ki})\right)\nonumber\\
\geq& \|P_{\spann\{\bt_k^*-\bt_i^*,\bt_k^*-\bt_j^*\}^\perp}(\bga_{ij})\|-\frac{\pi}{2}\sigma.\label{eq:tspan}
\end{align}

Denote $C^g_{ij}:=\{k\in C_{ij}: ki\in E_g, jk\in E_g\}$ so that $k\in C^g_{ij}$. Note that the underlying corruption model implies that $G(V,E_g)$ is an Erd\"{o}s-R\'{e}nyi graph $G(n,p(1-q))$. By combining  the assumption $np^2(1-q)^2>C_0\log n$ and  Lemma \ref{thm:cdist}, we obtain that the set of vertices $V$ is $C_1/\sqrt{\log n}$-well-distributed along $G(V, E_g)$ for some absolute constant $C_1$ with high probability. This fact and \eqref{eq:tspan} imply that with probability $1-n^{-5}$ 
\begin{align}
&\frac{1}{|C^g_{ij}|}\sum_{k\in C^g_{ij}}I_{AAB}(\bga_{ij};\bga_{jk},\bga_{ki})\nonumber\\
\geq &\frac{1}{|C^g_{ij}|}\sum_{k\in C^g_{ij}}\|P_{\spann\{\bt_k^*-\bt_i^*,\bt_k^*-\bt_j^*\}^\perp}(\bga_{ij})\|-\frac{\pi}{2}\sigma\nonumber\\
\geq &\frac{C_1}{\sqrt{\log n}}\|P_{(\bt_i^*-\bt_j^*)^\perp}\bga_{ij}\|-\frac{\pi}{2}\sigma=\frac{C_1}{\sqrt{\log n}}\|P_{\bga_{ij}^{*\perp}}\bga_{ij}\|-\frac{\pi}{2}\sigma\nonumber\\
\geq& \frac{C_1\pi}{2\sqrt{\log n}}\min(e_{ij}, \pi-e_{ij})-\frac{\pi}{2}\sigma.
\end{align}
\textbf{Case 2}. Either $jk\in E_g$ or $ki \in E_g$, but not both in $E_g$.\\
Let $Y^b_{ij}(k):=I_{AAB}(\bga_{ij};\bga_{jk},\bga_{ki})$. The arguments used for the estimates of case 2 of Section \ref{sec:eq1} and the fact that $f(x)\geq 0$ imply that $\mathbb{E}[Y^b_{ij}(k)]\geq 0$.\\
\textbf{Case 3}. $jk,ki\in E_b$\\
This case is exactly the same as case 3 of Section \ref{sec:eq1} and we thus use \eqref{eq:zg} for $\bz\sim U(S^2)$.

At last, combining the estimates of the 3 cases with respective probabilities $(1-q)^2$, $2q(1-q)$ and $q^2$ yields \eqref{eq:Eblower}.

\section{Experiments on Synthetic Data}\label{sec:syn}

\begin{figure}[!htbp]
\begin{center}
   \includegraphics[width=1\linewidth]{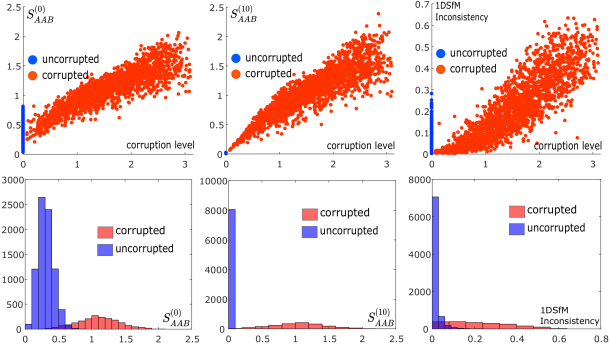}
\end{center}
   \caption{Demonstration of corruption identification for a synthetic dataset by naive AAB, IR-AAB and 1DSfM. The dataset was generated by the uniform corruption model UC($200,0.5,0.2,0$). The 3 columns of subfigures correspond to naive AAB, IR-AAB and 1DSfM respectively. The subfigures in the first row show the correlation of the computed statistics (on y-axis) with the corruption level (on x-axis).
   Edges with no corruption are blue and the rest are red. The subfigures in the second row are the histograms of computed statistics for both corrupted and uncorrupted edges.\label{fig:scatter}}
\end{figure}

We first illustrate the ability of the statistics obtained by naive AAB, IR-AAB and 1DSfM~\cite{1dsfm14} to separate corrupted and uncorrupted edges for a special synthetic dataset.
The dataset was randomly generated by the uniform corruption model with $n=200$, $p=0.5$, $q=0.2$ and $\sigma=0$. Figure \ref{fig:scatter} first shows the three statistics' values of edges as a function of their corruption levels. These corruption levels are measured by the angles of the corresponding pairwise directions with the uncorrupted pairwise directions. We first note that 1DSfM may assign zero values to corrupted edges, unlike naive AAB and IR-AAB, and has the largest variance per corruption level. We also note that IR-AAB assigns negligible values to uncorrupted points, unlike naive AAB and 1DSfM, and has the lowest variance at low corruption levels. The figure also shows the histograms of the statistics for both corrupted and uncorrupted points. Since the 1DSfM statistic (which is referred to in~\cite{1dsfm14} as inconsistency) obtains zero values for both corrupted and uncorrupted edges, it is hard to separate the whole histogram into two modes. On the other hand, naive AAB and IR-AAB can be nicely separated into two modes for this and other synthetic examples. For IR-AAB, but not naive AAB, this separation exactly recovers the uncorrupted edges in this particular example.

\begin{figure}[!htbp]
\begin{center}
   \includegraphics[width=1\linewidth]{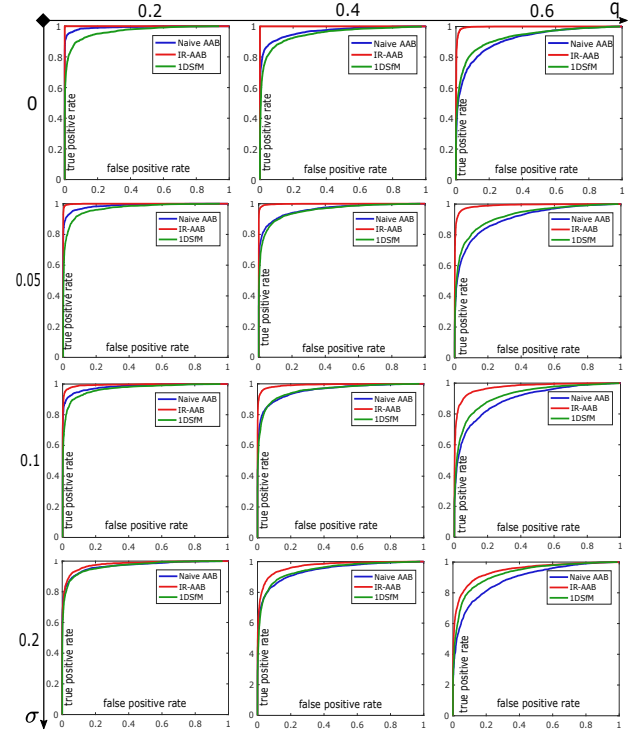}
\end{center}
   \caption{\label{fig:roc} ROC curves for corruption detection of naive AAB, IR-AAB and 1DSfM with varying corruption and noise levels.}
\end{figure}


Next we use ROC curves to diagnose the ability of naive AAB, IR-AAB and 1DSfM to detect corrupted edges in a similar synthetic data with varying percentages of corrupted edges and noise levels.  
The datasets were randomly generated by the uniform corruption model 
with $n=200$, $p=0.5$, $q=0.2$, $0.4$, $0.6$ and $\sigma=0$, $0.05$, $0.1$ and $0.2$.
For each statistic and choice of parameters, we assign 1000 equidistant thresholds between the largest and smallest values of this statistic, compute the true and false positive rates for recognizing uncorrupted points with values of this statistic above each threshold, and plot the corresponding ROC curve. We remark that the edges $ij\in E$ are recognized as corrupted when $\measuredangle(\bga_{ij},\bga_{ij}^*)>\sin^{-1}(\sigma)$. The resulting ROC curves are shown in Figure~\ref{fig:roc}, where a larger area under the ROC curve corresponds to better classification performance.

We note that classification based on IR-AAB consistently outperforms that of naive AAB and 1DSfM. Moreover, IR-AAB
has a clear advantage over naive AAB and 1DSfM at low and moderate noise levels ($\sigma=0$, $0.05$, $0.1$) among all levels of tested corruption. However, IR-AAB requires a certain portion of pairwise directions to be accurately estimated, and it thus does not significantly improve over the other two methods at high noise levels ($\sigma=0.2$). Naive AAB works well when the corruption and noise levels are relatively low. However, due to the misleading effect of corrupted neighboring edges, it may misclassify uncorrupted edges when the overall corruption or noise level is high ($q=0.6$ or $\sigma=0.2$). The performance of 1DSfM is not competitive. Indeed, it may frequently misclassify edges even at low corruption and noise levels, since it may converge to local extrema and also the 1D projection loses information.

\section{Experiments on Real Data}\label{sec:real}
We consider real datasets and compare the improvement obtained by preprocessing current camera location solvers with naive AAB, IR-AAB and 1DSfM. We use the 14 datasets from \cite{1dsfm14}. For each dataset, we exactly follow the pipeline suggested by \cite{cvprOzyesilS15} for estimating camera orientations and pairwise directions. Given the estimated pairwise directions from \cite{cvprOzyesilS15}, naive AAB, IR-AAB and 1DSfM are applied separately to delete $50\%$ of the edges with the highest corresponding statistics.
Different choices of $10\%$ and $90\%$ deleted edges are demonstrated in Appendix~\ref{ap:real}.
Since the graph may not be parallel rigid after deleting edges,  we extract its maximal parallel rigid component using a procedure suggested in \cite{KennedyDNT12}. We then apply to this component the following three different camera location solvers: LUD~\cite{cvprOzyesilS15} with IRLS implementation, CLS~\cite{TronV09_CLS1, TronV14_CLS2} with interior point method and ShapeFit~\cite{HandLV15} with ADMM implementation~\cite{GoldsteinHLVS16_shapekick}. We remark that although $50\%$ of edges are removed, the number of locations in the maximal parallel rigid graph is still close to the original graph.  For faster implementation of LUD, only a subset of the Piccadilly dataset with 500 locations is used.
For each dataset, each of the 3 statistics, and each of the 3 camera location solvers, we compute average and median distance (in meters) of the estimated camera locations to the ground truth locations\footnote{For each solver, the unknown scale and shift are estimated by least squares minimization with respect to the ground truth data.}. The latter ones are provided by \cite{1dsfm14}. The experimental results are recorded in Table \ref{tab:real}.

\begin{table*}[!htbp]
\centering \hspace{-0.1in}
\resizebox{2.1\columnwidth}{!}{
\renewcommand{\arraystretch}{1.05}
\tabcolsep=0.1cm
\begin{tabular}{|l||c|c||c|c||c|c||c|c||c|c||c|c||c|c||c|c||c|c||c|c||c|c||c|c|}
\hline
\multirow{2}{*}{Algorithms} & \multicolumn{8}{c||}{LUD~\cite{cvprOzyesilS15}} & \multicolumn{8}{c||}{CLS~\cite{TronV09_CLS1, TronV14_CLS2}} & \multicolumn{8}{c|}{ShapeFit~\cite{HandLV15}} \\
\text{} & \multicolumn{2}{c||}{None} & \multicolumn{2}{c||}{N-AAB} &  \multicolumn{2}{c||}{IR-AAB} & \multicolumn{2}{c||}{1DSfM}  &  \multicolumn{2}{c||}{None} & \multicolumn{2}{c||}{N-AAB} &  \multicolumn{2}{c||}{IR-AAB} & \multicolumn{2}{c||}{1DSfM}  &  \multicolumn{2}{c||}{None} & \multicolumn{2}{c||}{N-AAB} &  \multicolumn{2}{c||}{IR-AAB} & \multicolumn{2}{c|}{1DSfM} \\\hline
\text{Dataset}& {\large$\tilde{e}$} & {\large $\hat{e}$} & {\large$\tilde{e}$} & {\large $\hat{e}$} & {\large$\tilde{e}$} & {\large $\hat{e}$} & {\large$\tilde{e}$} & {\large $\hat{e}$} &{\large$\tilde{e}$} & {\large $\hat{e}$} & {\large$\tilde{e}$} & {\large $\hat{e}$} & {\large$\tilde{e}$} & {\large $\hat{e}$} & {\large$\tilde{e}$} & {\large $\hat{e}$} &{\large$\tilde{e}$} & {\large $\hat{e}$} & {\large$\tilde{e}$} & {\large $\hat{e}$} & {\large$\tilde{e}$} & {\large $\hat{e}$} & {\large$\tilde{e}$} & {\large $\hat{e}$} \\\hline
Alamo&$0.47$&$1.74$&$0.38$&$0.92$&$\mathbf{0.36}$&$\mathbf{0.85}$&$0.38$&$1.3$&$1.35$&$2.79$&$0.39$&$0.93$&$\mathbf{0.37}$&$\mathbf{0.69}$&$0.44$&$1.44$&$0.44$&$1.83$&$0.38$&$0.92$&$\mathbf{0.36}$&$\mathbf{0.85}$&$0.38$&$2.82$\\\hline
Madrid Metropolis & $1.84$&$5.94$&$1.28$&$3.57$&$\mathbf{1.21}$&$\mathbf{3.53}$&$1.46$&$5.79$&$7.1$&$11.2$&$1.48$&$5.28$&$\mathbf{1.26}$&$\mathbf{3.44}$&$2.73$&$3.59$&$14$&$27.3$&$1.51$&$17.8$&$\mathbf{1.22}$&$\mathbf{7.64}$&$4.61$&$29.58$\\\hline
Montreal N.D.&$0.56$&$1.22$&$0.4$&$0.61$&$\mathbf{0.39}$&$\mathbf{0.59}$&$0.53$&$1.2$&$0.9$&$1.79$&$\mathbf{0.4}$&$\mathbf{0.6}$&$0.41$&$\mathbf{0.6}$&$0.69$&$1.86$&$0.58$&$3.25$&$\mathbf{0.39}$&$0.63$&$\mathbf{0.39}$&$\mathbf{0.58}$&$0.61$&$4.08$\\\hline
Notre Dame &$0.29$&$0.85$&$0.26$&$0.6$&$\mathbf{0.24}$&$\mathbf{0.51}$&$0.28$&$1$&$1.05$&$2.12$&$0.36$&$0.86$&$\mathbf{0.27}$&$\mathbf{0.55}$&$0.61$&$1.47$&$0.24$&$0.96$&$0.23$&$0.58$&$\mathbf{0.22}$&$\mathbf{0.53}$&$0.24$&$1.27$\\\hline
NYC Library & $2.43$&$6.95$&$0.95$&$2.89$&$\mathbf{0.69}$&$\mathbf{2.24}$&$1.83$&$5.64$&$5.3$&$8.51$&$1.89$&$4.51$&$\mathbf{0.72}$&$\mathbf{2.52}$&$4.49$&$7.31$&$13.3$&$14.3$&$0.85$&$5.69$&$\mathbf{0.66}$&$\mathbf{2.23}$&$13.3$&$14.2$\\\hline
Piazza Del Popolo &$1.66$&$5.28$&$1.12$&$4.03$&$\mathbf{0.91}$&$\mathbf{1.54}$&$0.94$&$1.95$&$3.42$&$6.46$&$1.22$&$4.31$&$\mathbf{0.98}$&$\mathbf{1.57}$&$1.47$&$2.57$&$1.48$&$6.81$&$1.09$&$4.07$&$\mathbf{0.89}$&$\mathbf{1.51}$&$1$&$5.74$\\\hline
Piccadilly& $2.02$&$3.87$&$1.37$&$3.07$&$\mathbf{1.19}$&$\mathbf{2.69}$&$2.12$&$3.95$&$3.64$&$5.42$&$1.56$&$3.28$&$\mathbf{1.23}$&$\mathbf{2.42}$&$3.5$&$5.11$&$14.2$&$13.4$&$5.72$&$14.4$&$11.6$&$13.3$&$\mathbf{2.09}$&$\mathbf{6.39}$\\\hline
Roman Forum& $2.21$&$8.33$&$1.74$&$7.28$&$\mathbf{1.62}$&$\mathbf{7.13}$&$3.4$&$10.1$&$6.2$&$12.4$&$3.11$&$9.24$&$\mathbf{2.56}$&$\mathbf{8.58}$&$6.62$&$15.3$&$26.7$&$41$&$\mathbf{1.53}$&$\mathbf{12.7}$&$7.46$&$17.7$&$26.9$&$33.2$\\\hline
Tower of London& $4.03$&$17.9$&$2.41$&$4.79$&$\mathbf{2.33}$&$\mathbf{4.36}$&$2.83$&$15.8$&$16$&$27$&$2.6$&$4.87$&$\mathbf{2.36}$&$\mathbf{4.34}$&$12.6$&$24.8$&$2.41$&$16.9$&$2.34$&$4.74$&$\mathbf{2.27}$&$\mathbf{3.92}$&$2.48$&$20.1$\\\hline
Union Square& $7.57$&$11.7$&$\mathbf{7.24}$&$\mathbf{11.2}$&$7.3$&$11.3$&$7.89$&$12.9$&$8.03$&$12.5$&$\mathbf{7.39}$&$\mathbf{11.7}$&$7.84$&$\mathbf{11.7}$&$8.54$&$13.6$&$12.9$&$19$&$\mathbf{12.3}$&$\mathbf{18.6}$&$12.5$&$18.8$&$13.1$&$19.2$\\\hline
Vienna Cathedral& $7.26$&$13.1$&$6.86$&$14.9$&$\mathbf{4.21}$&$\mathbf{12.7}$&$9.05$&$17.4$&$9.59$&$\mathbf{13.7}$&$10$&$14.8$&$\mathbf{8.45}$&$13.9$&$8.62$&$15.6$&$28.6$&$36.6$&$28.5$&$36.5$&$28.5$&$\mathbf{36.4}$&$\mathbf{27.6}$&$\mathbf{36.4}$\\\hline
Yorkminster & $2.51$&$5.26$&$\mathbf{1.61}$&$6.74$&$1.62$&$4.91$&$2$&$\mathbf{3.69}$&$5.95$&$8.72$&$2.8$&$6.95$&$\mathbf{2.29}$&$\mathbf{6.36}$&$4.76$&$6.89$&$19.9$&$28.4$&$2.35$&$10.9$&$2.03$&$14.6$&$\mathbf{1.65}$&$\mathbf{4.51}$\\\hline
Ellis Island& $22$&$\mathbf{22.4}$&$23.5$&$23.7$&$24.7$&$24.6$&$\mathbf{21.7}$&$22.5$&$\mathbf{20.9}$&$\mathbf{22}$&$23.4$&$23.6$&$25.3$&$24.7$&$22.1$&$22.4$&$26.7$&$27.7$&$26.5$&$\mathbf{27.6}$&$26.6$&$27.8$&$\mathbf{26.4}$&$\mathbf{27.6}$\\\hline
Gendarmenmarkt&$17.5$&$\mathbf{38.8}$&$15.1$&$40.9$&$\mathbf{15}$&$41.3$&$17.1$&$40.6$&$20.7$&$\mathbf{40.9}$&$19.4$&$43.1$&$\mathbf{18.3}$&$42.1$&$19.2$&$42.3$&$32.8$&$51.7$&$32.7$&$52.1$&$32.5$&$52.1$&$\mathbf{32.4}$&$\mathbf{51.5}$\\\hline

\end{tabular}
}\vspace{0.05in}
\caption{Comparison of naive AAB, IR-AAB and 1DSfM for improving 3 location solvers (LUD, CLS, ShapeFit) using 14 datasets from~\cite{1dsfm14}. The median and mean distance from the estimated camera locations to the ground truth (provided in~\cite{1dsfm14}) are denoted by $\tilde{e}$ and $\hat{e}$ respectively. \label{tab:real}}
\end{table*}

These results 
show significant improvement of IR-AAB for all three camera location solvers. In particular, IR-AAB works best with LUD and CLS. For example, IR-AAB with LUD outperforms naive AAB and 1DSfM with LUD on 10 out of the 14 datasets in terms of both mean and median errors. For 2 additional datasets,  IR-AAB with LUD still improves over LUD.
For the two remaining datasets, Ellis Island and Gendarmenmarkt, which
contain highly inaccurate pairwise directions, none of the three statistics significantly improve any of the solvers.
We also note that while LUD is superior to CLS, after applying IR-AAB, both algorithms are comparable. Furthermore,  CLS with IR-AAB outperforms plain LUD.
We observe that 1DSfM outperforms IR-AAB on a few datasets when using ShapeFit. However, 1DSfM with ShapeFit is worse than plain ShapeFit on 6 other datasets. The inconsistent results of ShapeFit are due to its instability. Indeed, its formulation has a very weak constraint that cannot avoid collapsed solutions in the presence of highly corrupted pairwise directions and when, in particular, some locations have low degrees. On the other hand, both LUD and CLS have a very strong constraint, which avoids collapsed solutions.
Note that the preprocessing step results in a large component of the original graph with a possibly different topology than the original graph and thus ShapeFit may be more sensitive to the resulting subgraph, especially if it has some vertices with low degrees not present in the original graph. Due to this sensitivity, none of the preprocessing methods consistently outperforms the other ones when using ShapeFit. We remark that the instability of ShapeFit can be observed from the large variation of its estimation error using different outlier-removing methods and different datasets.

\begin{figure}[!htbp]
\begin{center}
   \includegraphics[width=1\linewidth]{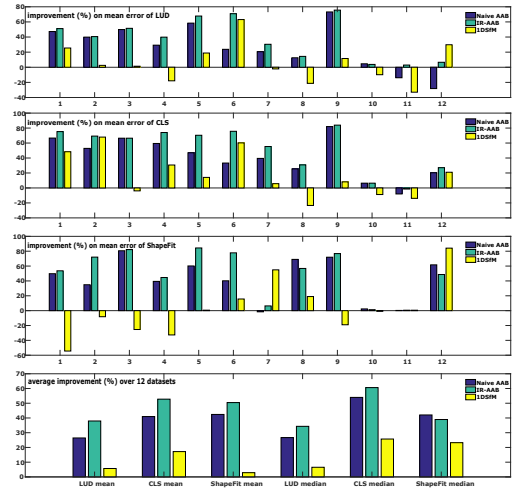}
\end{center}
   \caption{Percentage of improvement of location estimation by preprocessing 3 solvers with the 3 statistics. The first 3 subfigures illustrate the mean error improvement for LUD, CLS and ShapeFit respectively. Numbers 1-12 on the horizontal axis are the indices of the first 12 datasets in Table  \ref{tab:real}. Negative number of improvement rate corresponds to increase of the estimation error after removing edges. The last subfigure illustrates the averaged mean and median improvement of the 3 statistics over the first 12 datasets when preprocessing the three solvers by the three statistics. \label{fig:improve}}
\end{figure}

Figure \ref{fig:improve} illustrates the improvement of the three preprocessing algorithms (naive AAB, IR-AAB and 1DSfM) over the three solvers (LUD, CLS and ShapeFit). The improvement is measured by the following formula:
\begin{equation}
\text{Improvement} = \frac{e_{\text{before}}-e_{\text{after}}}{e_{\text{before}}}\cdot 100\%,
\end{equation}
where $e_{\text{before}}$ is the mean/median error of estimated camera locations on the whole graph by the given solver without preprocessing and $e_{\text{after}}$ is the mean/median error of the same solver after removing $50\%$ of edges by the given preprocessing algorithm. The two datasets with highly inaccurate pairwise directions, Ellis Island and Gendarmenmarkt, were removed. The first three subfigures indicate results of mean error for each solver separately and the last subfigure demonstrates the averaged mean and median errors result among the 12 remaining datasets.
It is evident that IR-AAB has the best overall performance in improving the three solvers. On the other hand, 1DSfM has the worst performance. For example, IR-AAB succeeds in improving LUD's performance with average mean-error rate of $38\%$ and it consistently reduces the estimation error of LUD on all of these datasets. On the other hand, 1DSfM has average mean-error improvement rate for LUD of $5.7\%$, whereas on five datasets preprocessing by 1DSfM increases the estimation error of LUD. For comparison, naive AAB has average mean-error improvement rate for LUD of $26.5\%$, whereas on two datasets preprocessing by naive AAB increases the estimation error of LUD. For all of the 3 statistics, the overall improvement of preprocessing with CLS is more significant than preprocessing with LUD and ShapeFit. Indeed the averaged mean and median improvement rates of CLS when preprocessing by IR-AAB is more than 50\%.
This is not surprising as CLS is not robust to corruption.
For ShapeFit, the average improvements over the mean errors when preprocessing with naive AAB, IR-AAB and 1DSfM are $42.4\%$, $50.4\%$ and $2.9\%$ respectively. However, when considering each individual dataset, IR-AAB and naive AAB may not consistently outperform 1DSfM due to the instability of ShapeFit discussed above.

We report the computational speed of the algorithms on the largest dataset: Roman Forum, which has 967 locations. While Piccadilly has 2226 locations, it was run with 500 locations to ease the computational time for LUD and for extracting the maximal parallel rigid component. The computations were performed on a machine with 2.5GHz Intel i5 quad core processors and 8GB memory.
The total time needed to compute 1DSfM, naive AAB and IR-AAB is 2, 5 and 8 seconds respectively.
For comparison, the total time to run CLS, ShapeFit and LUD is 8, 8 and 160 seconds respectively. We expect the runtime of ADMM for LUD to be comparable to that of ShapeFit.
The slowest component was finding the maximal parallel rigid component. For Roman Forum, it took 550 seconds, while it took less than 200 seconds for the other datasets.

\section{Conclusion} \label{sec:conc}
We proposed the AAB statistic for estimating the underlying corruption level on camera pairwise directions.
We improved this estimation by incorporating a careful reweighting strategy.
We further established theoretical guarantee on the accuracy of the non-reweighted statistic, i.e., naive AAB, for detecting corrupted edges when the corruption and noise levels are sufficiently low. The experiments on both synthetic data and real data show the significant advantage of applying the reweighting strategy with the AAB statistic. Applying our method as a preprocessing step significantly improves the performance of current camera location solvers.

This work suggests several interesting future projects. First of all, we believe that a similar strategy can be developed for improving camera orientation estimation. Second of all, we are interested in theoretically guaranteeing the reweighting strategy for segmenting corrupted and uncorrupted edges. Third of all, an interesting direction for future work is to study and provide guarantees for synthetic models that more realistically mirror real scenarios. At last, we find it important to develop a faster method for extracting the maximal parallel rigid graph so that the total runtime can be significantly reduced.

\section*{Acknowledgement}
This work was supported by NSF award DMS-14-18386. We thank Soumyadip Sengupta, Thomas Goldstein, Tal Amir and Paul Hand for providing codes and real data. We also thank the anonymous reviewers and Tyler Maunu for helpful comments on an earlier version of this manuscript.

{\small
\bibliographystyle{ieee}
\bibliography{egbib}
}

\newpage
\appendix
\section{Appendix}
\subsection{Proof of the AAB Formula \eqref{eq:aabf}}\label{ap:aabf}
As is shown in Figure \ref{fig:s2}, $\Omega(\bga_1,\bga_2)$ is exactly the shortest path on the manifold $S^2$ between $-\bga_1$ and $-\bga_2$. Since $I_{AAB}(\bga_3;\bga_1,\bga_2)$ is the length of shortest path between $\bga_3$ and $\Omega(\bga_1,\bga_2)$, it can be computed via the following procedure:
Let $\bga_p$ be the orthogonal projection of $\bga_3$ onto $\spann\{\bga_1,\bga_2\}$, then
\begin{multline}
I_{AAB}(\bga_3;\bga_1,\bga_2)\\
=\begin{cases}
\measuredangle(\bga_p,\bga_3), &\text{if }\frac{\bga_p}{\|\bga_p\|}\in \Omega(\bga_1,\bga_2);\\
\min\left(\measuredangle(\bga_1,\bga_3),\measuredangle(\bga_2,\bga_3)\right), &\text{otherwise}.
\end{cases}
\end{multline}

By the definition of $\bga_p$ it can be expressed as $\lambda_1\bga_1+\lambda_2\bga_2$, where $(\bga_3-\lambda_1\bga_1-\lambda_2\bga_2)$$\perp \spann\{\bga_1,\bga_2\}$. That is, $\langle \bga_3-\lambda_1\bga_1-\lambda_2\bga_2,\bga_1\rangle = \langle \bga_3-\lambda_1\bga_1-\lambda_2\bga_2,\bga_2\rangle=0$. Thus, we obtain the following system of equations for $\lambda_1$ and $\lambda_2$
\begin{eqnarray}
\label{eq:lambda}
&\lambda_1+z\lambda_2=x  \\
&z\lambda_1+\lambda_2=y,
\end{eqnarray}
where we recall that $x=\bga_1^T\bga_3$, $y=\bga_2^T\bga_3$ and $z=\bga_1^T\bga_2$.
The solution of~\eqref{eq:lambda} is given by $\lambda_1=(x-yz)/(1-z^2)$,
 $\lambda_2=(y-xz)/(1-z^2)$. Note that $\bga_p/\|\bga_p\|\in \Omega(\bga_1,\bga_2)$ if and only if $\lambda_1<0$ and $\lambda_2<0$. That is, when $y<xz$ and $x<yz$,
 \begin{multline}
 I_{AAB}(\bga_3;\bga_1,\bga_2)=\cos^{-1}(\bga_p^T\bga_3) \\
 =\cos^{-1}(\lambda_1\bga_1^T\bga_3+\lambda_2\bga_2^T\bga_3)\\
 =\cos^{-1}(\lambda_1x+\lambda_2y)
=\frac{x^2+y^2-2xyz}{1-z^2}.
\end{multline}
Otherwise,
\begin{multline}
 I_{AAB}(\bga_3;\bga_1,\bga_2)=\min\left(\measuredangle(\bga_1,\bga_3),\measuredangle(\bga_2,\bga_3)\right)\\
 =\cos^{-1}\left(\max\left(\bga_1^T\bga_3,\bga_2^T\bga_3\right)\right)
 =\cos^{-1}\left(\max\left(x,y\right)\right).
\end{multline}
This concludes the proof of formula \eqref{eq:aabf}.
\subsection{Proof of Lemma \ref{thm:f}}
\label{ap:fx}
\begin{proof}
Let $l(x_1,x_2)$ denote the shortest path on $S^2$ connecting the points $x_1$ and $x_2$. Let $\bu_1=-\bv_1, \bu=-\bv$. Note that $x=\measuredangle(\bv_2(x),\bu_1)$ by the definition of  $\bv_2(x)$ and $\bu_1$.
\begin{align}
f(x)=&\mathbb{E}[\min_{\by\in l(\bu_1,\bu)}d_g(\bv_2(x),\by)|\bu\sim U(S^2)]\nonumber\\
=&\int\min_{\by\in l(\bu_1,\bu)}d_g(\bv_2(x),\by)p(\bu)d\bu\nonumber\\
=&\int\limits_{d_g(\bu,\bu_1)\leq x}\min_{\by\in l(\bu_1,\bu)}d_g(\bv_2(x),\by)p(\bu)d\bu\nonumber\\
&+\int\limits_{d_g(\bu,\bu_1)> x}\min_{\by\in l(\bu_1,\bu)}d_g(\bv_2(x),\by)p(\bu)d\bu\nonumber\\
=&\int\limits_{d_g(\bu,\bu_1)\leq x}d_g(\bu,\bu_1)p(\bu)d\bu
+\int\limits_{d_g(\bu,\bu_1)> x}x p(\bu)d\bu\nonumber\\
=&\frac{1}{4\pi}\left[\int_0^{2\pi}\int_0^x\sin\phi\cdot \phi\, d\phi d\theta+\int_0^{2\pi}\int_x^\pi\sin\phi\cdot x\, d\phi d\theta\right]\nonumber\\
=& \frac{1}{2}(x+\sin x),
\label{eq:fx}
\end{align}
where $\theta$ and $\phi$ are azimuthal angle and polar angle in spherical coordinate system respectively.
\end{proof}


\subsection{Additional Real Data Experiments}\label{ap:real}
Table \ref{tab:2} and  \ref{tab:3} are similar to Table \ref{tab:real} of Section \ref{sec:real}, however, while in Table \ref{tab:real} $50\%$ of edges were removed, 
in the new tables $10\%$ and $90\%$ of edges are removed.
\begin{table*}[!htbp]
\centering \hspace{-0.1in}
\resizebox{2.1\columnwidth}{!}{
\renewcommand{\arraystretch}{1.05}
\tabcolsep=0.1cm
\begin{tabular}{|l||c|c||c|c||c|c||c|c||c|c||c|c||c|c||c|c||c|c||c|c||c|c||c|c|}
\hline
\multirow{2}{*}{Algorithms} & \multicolumn{8}{c||}{LUD\cite{cvprOzyesilS15}} & \multicolumn{8}{c||}{CLS~\cite{TronV09_CLS1, TronV14_CLS2}} & \multicolumn{8}{c|}{ShapeFit~\cite{HandLV15}} \\ 
\text{} & \multicolumn{2}{c||}{None} & \multicolumn{2}{c||}{N-AAB} &  \multicolumn{2}{c||}{IR-AAB} & \multicolumn{2}{c||}{1DSfM}  &  \multicolumn{2}{c||}{None} & \multicolumn{2}{c||}{N-AAB} &  \multicolumn{2}{c||}{IR-AAB} & \multicolumn{2}{c||}{1DSfM}  &  \multicolumn{2}{c||}{None} & \multicolumn{2}{c||}{N-AAB} &  \multicolumn{2}{c||}{IR-AAB} & \multicolumn{2}{c|}{1DSfM}  \\\hline
\text{Dataset}& {\large$\tilde{e}$} & {\large $\hat{e}$} & {\large$\tilde{e}$} & {\large $\hat{e}$} & {\large$\tilde{e}$} & {\large $\hat{e}$} & {\large$\tilde{e}$} & {\large $\hat{e}$} &{\large$\tilde{e}$} & {\large $\hat{e}$} & {\large$\tilde{e}$} & {\large $\hat{e}$} & {\large$\tilde{e}$} & {\large $\hat{e}$} & {\large$\tilde{e}$} & {\large $\hat{e}$} &{\large$\tilde{e}$} & {\large $\hat{e}$} & {\large$\tilde{e}$} & {\large $\hat{e}$} & {\large$\tilde{e}$} & {\large $\hat{e}$} & {\large$\tilde{e}$} & {\large $\hat{e}$} \\\hline
Alamo&$0.47$&$1.74$&$0.43$&$1.26$&$\mathbf{0.38}$&$\mathbf{1.06}$&$0.45$&$1.84$&$1.35$&$2.79$&$0.52$&$1.43$&$\mathbf{0.4}$&$\mathbf{1.2}$&$0.71$&$2.2$&$0.44$&$1.83$&$0.42$&$2.57$&$\mathbf{0.39}$&$\mathbf{1.54}$&$0.44$&$2.04$\\\hline
Madrid Metropolis & $1.84$&$5.94$&$\mathbf{1.66}$&$4.72$&$\mathbf{1.66}$&$\mathbf{4.47}$&$1.68$&$5.46$&$7.1$&$11.2$&$4.16$&$7.72$&$\mathbf{3.68}$&$\mathbf{7.06}$&$4.45$&$9.08$&$14$&$27.3$&$1.49$&$9.06$&$\mathbf{1.45}$&$\mathbf{5.44}$&$1.47$&$10.92$\\\hline
Montreal N.D.&$0.56$&$1.22$&$\mathbf{0.48}$&$0.81$&$0.49$&$\mathbf{0.75}$&$0.56$&$1.29$&$0.9$&$1.79$&$\mathbf{0.49}$&$0.8$&$0.51$&$\mathbf{0.76}$&$0.68$&$1.65$&$0.58$&$3.25$&$\mathbf{0.46}$&$0.83$&$\mathbf{0.46}$&$\mathbf{0.78}$&$0.65$&$3.66$\\\hline
Notre Dame &$0.29$&$0.85$&$0.28$&$0.79$&$\mathbf{0.27}$&$0.81$&$0.28$&$\mathbf{0.78}$&$1.05$&$2.12$&$0.6$&$1.28$&$\mathbf{0.53}$&$\mathbf{1.25}$&$0.73$&$1.36$&$0.24$&$0.96$&$\mathbf{0.23}$&$\mathbf{0.69}$&$0.24$&$0.73$&$\mathbf{0.23}$&$0.7$\\\hline
NYC Library& $2.43$&$6.95$&$1.84$&$6.3$&$\mathbf{1.62}$&$5.29$&$1.86$&$\mathbf{5.03}$&$5.3$&$8.51$&$4.33$&$7.93$&$\mathbf{3.88}$&$\mathbf{6.93}$&$4.76$&$7.34$&$13.2$&$14.2$&$13.1$&$14.1$&$\mathbf{13}$&$\mathbf{13.9}$&$13.3$&$14.3$\\\hline
Piazza Del Popolo &$1.66$&$5.28$&$1.42$&$\mathbf{5.23}$&$\mathbf{1.41}$&$5.47$&$1.51$&$5.34$&$3.42$&$6.46$&$2.2$&$6.16$&$\mathbf{1.84}$&$\mathbf{6.08}$&$2.56$&$6.14$&$1.47$&$6.81$&$\mathbf{1.31}$&$\mathbf{5.95}$&$1.35$&$6.76$&$1.42$&$6.75$\\\hline
Piccadilly& $2.02$&$3.87$&$1.79$&$3.45$&$\mathbf{1.64}$&$\mathbf{3.29}$&$1.85$&$3.62$&$3.64$&$5.42$&$2.89$&$4.46$&$\mathbf{2.82}$&$\mathbf{4.29}$&$3.37$&$4.98$&$13.4$&$14.2$&$1.4$&$4.95$&$\mathbf{1.35}$&$\mathbf{4.35}$&$13.4$&$14.1$\\\hline
Roman Forum& $2.21$&$8.33$&$1.84$&$7.86$&$\mathbf{1.77}$&$\mathbf{7.61}$&$2.18$&$8.74$&$6.2$&$12.4$&$3.49$&$9.3$&$\mathbf{4.37}$&$\mathbf{8.94}$&$6.23$&$12.2$&$26.7$&$41$&$21.4$&$30.6$&$\mathbf{12.1}$&$\mathbf{19.5}$&$15.1$&$39.3$\\\hline
Tower of London& $4.03$&$17.9$&$2.74$&$15.9$&$\mathbf{2.67}$&$\mathbf{8.85}$&$3.26$&$17.5$&$16$&$27$&$5.87$&$17.5$&$\mathbf{2.78}$&$\mathbf{9.2}$&$15.3$&$26.6$&$\mathbf{2.41}$&$\mathbf{16.9}$&$2.49$&$19.5$&$2.76$&$31.4$&$2.48$&$17.4$\\\hline
Union Square& $7.57$&$11.7$&$\mathbf{7.29}$&$\mathbf{11.2}$&$7.5$&$11.8$&$7.97$&$12.3$&$8.03$&$12.5$&$\mathbf{7.82}$&$\mathbf{12.1}$&$8.06$&$12.6$&$8.6$&$13.1$&$12.9$&$19$&$\mathbf{12.7}$&$19$&$12.8$&$19.2$&$12.8$&$19$\\\hline
Vienna Cathedral& $7.26$&$13.1$&$6.41$&$13.4$&$6.6$&$13.9$&$\mathbf{5.68}$&$\mathbf{11.7}$&$9.59$&$13.7$&$9.4$&$13.4$&$9.62$&$13.9$&$\mathbf{7.36}$&$\mathbf{11.4}$&$\mathbf{28.6}$&$36.6$&$28.7$&$36.7$&$29.8$&$\mathbf{35.9}$&$28.7$&$36.6$\\\hline
Yorkminster & $2.51$&$5.26$&$1.73$&$\mathbf{4.32}$&$\mathbf{1.7}$&$4.63$&$2.05$&$4.72$&$5.95$&$8.72$&$3.61$&$\mathbf{6.1}$&$\mathbf{3.44}$&$6.33$&$5.87$&$8.46$&$19.9$&$28.4$&$1.66$&$15.6$&$\mathbf{1.56}$&$\mathbf{12.5}$&$14.7$&$16.8$\\\hline
Ellis Island& $\mathbf{22}$&$\mathbf{22.4}$&$22.6$&$23.2$&$23.8$&$23.6$&$22.2$&$22.8$&$\mathbf{20.9}$&$\mathbf{22}$&$22.6$&$23.3$&$24.3$&$23.7$&$22.5$&$22.8$&$26.7$&$27.7$&$26.6$&$\mathbf{27.5}$&$\mathbf{26.5}$&$27.8$&$26.7$&$27.7$\\\hline
Gendarmenmarkt&$17.5$&$\mathbf{38.8}$&$16.6$&$38.9$&$16.7$&$39.1$&$\mathbf{16.5}$&$38.9$&$20.7$&$40.9$&$18.5$&$41.3$&$\mathbf{17.8}$&$\mathbf{40.8}$&$18.7$&$\mathbf{40.8}$&$32.8$&$51.6$&$32.8$&$51.7$&$32.9$&$51.8$&$32.8$&$51.6$\\\hline

\end{tabular}
}\vspace{0.05in}
\caption{Comparison of naive AAB, IR-AAB and 1DSfM for improving 3 location solvers (LUD, CLS, ShapeFit) using 14 datasets from~\cite{1dsfm14}. Using any of the three statistics, {\color{blue}$10\%$} of edges are removed. The median and mean distance from the estimated camera locations to the ground truth (provided in~\cite{1dsfm14}) are denoted by $\tilde{e}$ and $\hat{e}$ respectively.}\label{tab:2}
\end{table*}

\begin{table*}[!htbp]
\centering \hspace{-0.1in}
\resizebox{2.1\columnwidth}{!}{
\renewcommand{\arraystretch}{1.05}
\tabcolsep=0.1cm
\begin{tabular}{|l||c|c||c|c||c|c||c|c||c|c||c|c||c|c||c|c||c|c||c|c||c|c||c|c|}
\hline
\multirow{2}{*}{Algorithms} & \multicolumn{8}{c||}{LUD\cite{cvprOzyesilS15}} & \multicolumn{8}{c||}{CLS~\cite{TronV09_CLS1, TronV14_CLS2}} & \multicolumn{8}{c|}{ShapeFit~\cite{HandLV15}} \\ 
\text{} & \multicolumn{2}{c||}{None} & \multicolumn{2}{c||}{N-AAB} &  \multicolumn{2}{c||}{IR-AAB} & \multicolumn{2}{c||}{1DSfM}  &  \multicolumn{2}{c||}{None} & \multicolumn{2}{c||}{N-AAB} &  \multicolumn{2}{c||}{IR-AAB} & \multicolumn{2}{c||}{1DSfM}  &  \multicolumn{2}{c||}{None} & \multicolumn{2}{c||}{N-AAB} &  \multicolumn{2}{c||}{IR-AAB} & \multicolumn{2}{c|}{1DSfM}  \\\hline
\text{Dataset}& {\large$\tilde{e}$} & {\large $\hat{e}$} & {\large$\tilde{e}$} & {\large $\hat{e}$} & {\large$\tilde{e}$} & {\large $\hat{e}$} & {\large$\tilde{e}$} & {\large $\hat{e}$} &{\large$\tilde{e}$} & {\large $\hat{e}$} & {\large$\tilde{e}$} & {\large $\hat{e}$} & {\large$\tilde{e}$} & {\large $\hat{e}$} & {\large$\tilde{e}$} & {\large $\hat{e}$} &{\large$\tilde{e}$} & {\large $\hat{e}$} & {\large$\tilde{e}$} & {\large $\hat{e}$} & {\large$\tilde{e}$} & {\large $\hat{e}$} & {\large$\tilde{e}$} & {\large $\hat{e}$} \\\hline
Alamo&$0.47$&$1.74$&$0.37$&$0.81$&$\mathbf{0.35}$&$\mathbf{0.59}$&$0.37$&$0.93$&$1.35$&$2.79$&$0.36$&$0.8$&$\mathbf{0.35}$&$\mathbf{0.6}$&$0.42$&$0.98$&$0.44$&$1.83$&$0.36$&$0.82$&$\mathbf{0.35}$&$\mathbf{0.72}$&$0.37$&$0.91$\\\hline
Madrid Metropolis & $1.84$&$5.94$&$1.06$&$2.47$&$\mathbf{0.98}$&$\mathbf{2.43}$&$1.39$&$4.86$&$7.1$&$11.2$&$1.26$&$2.85$&$\mathbf{1.12}$&$\mathbf{2.72}$&$2.17$&$6.26$&$14$&$27.3$&$\mathbf{3.03}$&$\mathbf{6.78}$&$4.7$&$14$&$21.8$&$32.3$\\\hline
Montreal N.D.&$0.56$&$1.22$&$0.38$&$0.57$&$\mathbf{0.37}$&$\mathbf{0.56}$&NA&NA&$0.9$&$1.79$&$0.4$&$0.59$&$\mathbf{0.37}$&$\mathbf{0.55}$&NA&NA&$0.58$&$3.25$&$0.39$&$\mathbf{0.57}$&$\mathbf{0.37}$&$0.58$&NA&NA\\\hline
Notre Dame &$0.29$&$0.85$&$0.23$&$0.47$&$\mathbf{0.2}$&$\mathbf{0.38}$&$0.27$&$0.74$&$1.05$&$2.12$&$0.27$&$0.57$&$\mathbf{0.21}$&$\mathbf{0.43}$&$0.66$&$1.23$&$\mathbf{0.24}$&$\mathbf{0.96}$&$0.28$&$1.66$&$0.29$&$1.14$&$\mathbf{0.24}$&$1.32$\\\hline
NYC Library& $2.43$&$6.95$&$0.81$&$3.95$&$\mathbf{0.61}$&$\mathbf{1.37}$&NA&NA&$5.3$&$8.51$&$0.8$&$2.36$&$\mathbf{0.63}$&$\mathbf{1.49}$&NA&NA&$13.3$&$14.3$&$1.4$&$8.16$&$\mathbf{0.7}$&$\mathbf{2.76}$&NA&NA\\\hline

Piazza Del Popolo &$1.66$&$5.28$&$0.93$&$1.55$&$\mathbf{0.75}$&$\mathbf{1.28}$&$0.96$&$2.1$&$3.42$&$6.46$&$0.86$&$1.42$&$\mathbf{0.82}$&$\mathbf{1.33}$&$1.38$&$2.56$&$1.48$&$6.81$&$0.94$&$3.56$&$\mathbf{0.78}$&$\mathbf{1.33}$&$0.9$&$1.95$\\\hline

Piccadilly& $2.02$&$3.87$&$1.24$&$2.31$&$\mathbf{0.9}$&$\mathbf{2.04}$&$2.79$&$4.62$&$3.64$&$5.42$&$1.21$&$2.15$&$\mathbf{0.97}$&$\mathbf{2.03}$&$2.88$&$4.54$&$13.4$&$14.2$&$7.04$&$12$&$\mathbf{1.11}$&$\mathbf{5.9}$&$8.99$&$13$\\\hline

Roman Forum& $2.21$&$8.33$&$1.47$&$5.02$&$\mathbf{1.15}$&$\mathbf{3.66}$&$4.1$&$13.9$&$6.2$&$12.4$&$1.88$&$\mathbf{5.38}$&$\mathbf{1.44}$&$5.44$&$8.69$&$17.2$&$26.7$&$41$&$15.8$&$31.1$&$\mathbf{5.1}$&$\mathbf{21.1}$&$15.5$&$39.9$\\\hline

Tower of London& $4.03$&$17.9$&$\mathbf{2.39}$&$3.68$&$2.4$&$\mathbf{3.49}$&$2.78$&$14.34$&$16$&$27$&$2.45$&$\mathbf{4.13}$&$\mathbf{2.26}$&$4.2$&$9.58$&$20.4$&$\mathbf{2.41}$&$16.9$&$2.62$&$\mathbf{6.47}$&$2.6$&$6.86$&$5.3$&$56.5$\\\hline

Union Square& $7.57$&$11.7$&$5.96$&$9.84$&$6.37$&$11.5$&$\mathbf{5.73}$&$\mathbf{9.04}$&$8.03$&$12.5$&$\mathbf{5.91}$&$\mathbf{9.15}$&$10.2$&$16.5$&$6.11$&$9.49$&$12.9$&$19$&$12.7$&$16.3$&$13$&$17.6$&$\mathbf{11.7}$&$\mathbf{14.1}$\\\hline

Vienna Cathedral& $7.26$&$13.1$&$3.69$&$8.88$&$\mathbf{2.41}$&$\mathbf{8.71}$&$8.99$&$17.4$&$9.59$&$13.7$&$8.16$&$12$&$\mathbf{4.65}$&$\mathbf{10.8}$&$9.48$&$19.1$&$28.6$&$36.6$&$28.9$&$35.9$&$\mathbf{2.12}$&$\mathbf{9.11}$&$24.3$&$33.8$\\\hline

Yorkminster & $2.51$&$5.26$&$1.4$&$3$&$\mathbf{1.26}$&$\mathbf{2.7}$&$1.8$&$3.98$&$5.95$&$8.72$&$2.72$&$4.46$&$\mathbf{1.44}$&$\mathbf{2.82}$&$3.54$&$5.53$&$19.9$&$28.4$&$3.55$&$18.4$&$\mathbf{1.75}$&$\mathbf{4.75}$&$2.75$&$6.66$\\\hline

Ellis Island& $\mathbf{22}$&$\mathbf{22.4}$&$22.1$&$23.2$&$25.6$&$25.3$&$24.3$&$24.4$&$\mathbf{20.9}$&$22$&$23$&$23.5$&$26.1$&$25$&$21$&$\mathbf{21.9}$&$26.7$&$27.7$&$26.3$&$\mathbf{27.2}$&$\mathbf{26.2}$&$27.3$&$\mathbf{26.2}$&$27.3$\\\hline
Gendarmenmarkt & $\mathbf{17.5}$&$\mathbf{38.8}$&$20.9$&$46.1$&$34$&$61.3$&$17.7$&$40.2$&$20.7$&$\mathbf{40.9}$&$22$&$47.3$&$33.2$&$61.7$&$\mathbf{19.3}$&$42.7$&$32.8$&$51.7$&$33.2$&$55$&$\mathbf{32.5}$&$62.7$&$33.1$&$\mathbf{51.5}$\\\hline
\end{tabular}
}\vspace{0.05in}
\caption{Comparison of naive AAB, IR-AAB and 1DSfM for improving 3 location solvers (LUD, CLS, ShapeFit) using 14 datasets from~\cite{1dsfm14}. Using any of the three statistics, {\color{blue}$90\%$} of edges are removed. The median and mean distance from the estimated camera locations to the ground truth (provided in~\cite{1dsfm14}) are denoted by $\tilde{e}$ and $\hat{e}$ respectively. Even after removing $90\%$ of edges, in most of the cases the maximal parallel rigid subgraph still contains $>50\%$ camera locations. ``NA" means that the resulting maximal parallel rigid component had only $16$ or less locations, whereas in the rest of cases the maximal parallel rigid component had at least 100 locations.}\label{tab:3}
\end{table*}

\end{document}